\documentclass[letterpaper]{article} 
\usepackage{aaai25}  
\usepackage{times}  
\usepackage{helvet}  
\usepackage{courier}  
\usepackage[hyphens]{url}  
\usepackage{graphicx} 
\usepackage{natbib}  
\usepackage{caption} 
\usepackage{algorithm}
\usepackage{algorithmic}

\usepackage{newfloat}
\usepackage{listings}
\DeclareCaptionStyle{ruled}{labelfont=normalfont,labelsep=colon,strut=off} 
\floatstyle{ruled}
\newfloat{listing}{tb}{lst}{}
\floatname{listing}{Listing}
\newcommand{\sysname}{{RCIR}}
\newcommand{\modulea}{{\textbf{Adapter}}}
\newcommand{\moduleb}
{{\textbf{Controller}}}
\newcommand*\sysh{\includegraphics[width=8pt]{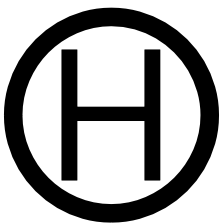}}
\newcommand*\sysr{\includegraphics[width=8pt]{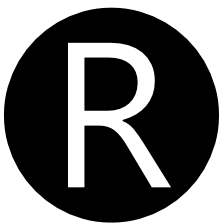}}

\usepackage{dsfont}
\usepackage{amsthm}
\usepackage{amsmath}
\newtheorem{theorem}{Theorem}
\newtheorem{lemma}{Lemma}
\usepackage{caption}
\usepackage{subcaption}
\usepackage{bm}
\usepackage{xcolor}
\definecolor{color_wrong}{HTML}{FE1E1F} 
\definecolor{color_correct}{HTML}{0AAF54} 
\definecolor{color_01}{HTML}{D65DB1} 
\definecolor{color_02}{HTML}{2B69AB} 

\usepackage{cleveref}
\crefname{figure}{Fig.}{Figs.}
\crefname{equation}{Eq.}{Eqs.}
\crefname{section}{Sec.}{Secs.}
\crefname{lemma}{Lemma}{lemmas}
\crefname{theorem}{Theorem}{theorems}
\crefname{algorithm}{Alg.}{algorithms}

\title{Risk Controlled Image Retrieval}
\author{
    Kaiwen Cai$^1$, Chris Xiaoxuan Lu$^2$, Xingyu Zhao$^3$, Wei Huang$^4$, Xiaowei Huang$^1$\footnote{corresponding author: xiaowei.huang@liverpool.ac.uk}\\
}
\affiliations{
    \textsuperscript{\rm 1}University of Liverpool,
    \textsuperscript{\rm 2}University College London, 
    \textsuperscript{\rm 3}University of Warwick,
    \textsuperscript{\rm 4}Purple Mountain Laboratories\\
}

\begin{document}

\maketitle

\begin{abstract}
Most image retrieval research prioritizes improving predictive performance, often overlooking situations where the reliability of predictions is equally important.
The gap between model performance and reliability requirements highlights the need for a systematic approach to analyze and address the risks associated with image retrieval.
Uncertainty quantification technique can be applied to mitigate this issue by assessing uncertainty for retrieval sets, but it 
provides only a heuristic estimate of uncertainty rather than a \textit{guarantee}.
To address these limitations, we present Risk Controlled Image Retrieval (\sysname), which generates retrieval sets with coverage guarantee, i.e., retrieval sets that are guaranteed to contain the true nearest neighbors with a predefined probability. 
\sysname\ can be easily integrated with existing uncertainty-aware image retrieval systems, agnostic to data distribution and model selection. 
To the best of our knowledge, this is the first work that provides coverage guarantees to image retrieval. The validity and efficiency of \sysname\ are demonstrated on four real-world datasets: CAR-196, CUB-200, Pittsburgh, and ChestX-Det.

\end{abstract}

\section{Introduction}
\begin{figure}[t]
    \centering
    \includegraphics[width=1\columnwidth]{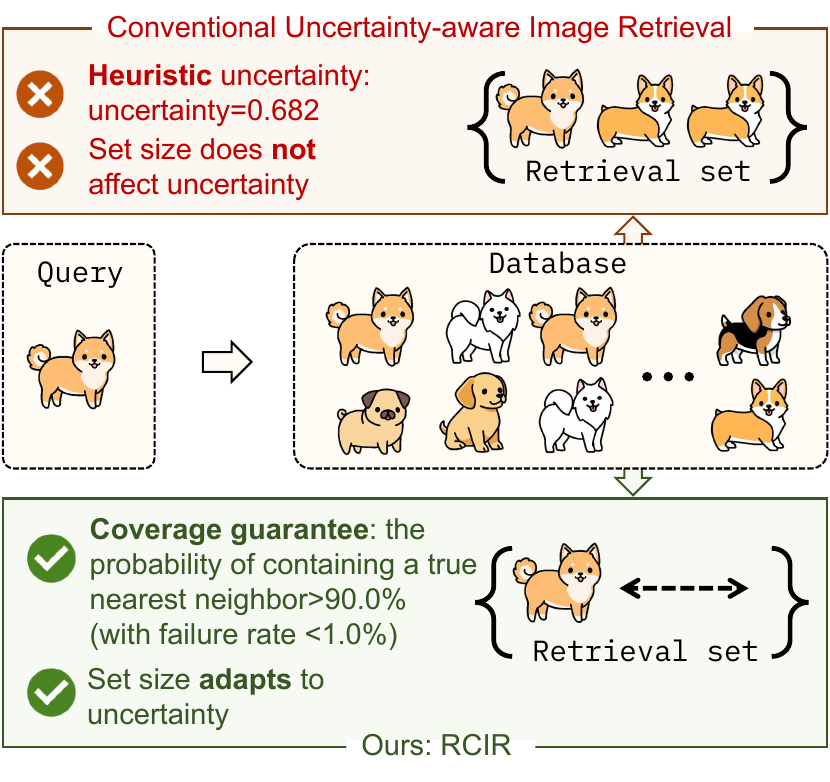} 
    \caption{Ilustration of our risk-controlled image retrieval \sysname\ and conventional uncertainty-aware image retrieval systems (e.g., \cite{warburg2021bayesian}): An uncertainty-aware image retrieval provides a heuristic uncertainty, and has a fixed retrieval set size. In contrast, \sysname\ provides a retrieval set that is guaranteed to cover a true nearest neighbor with a user-specified risk requirement (probability of containing, and failure rate), and moreover the retrieval size adapts to the uncertainty of the query sample and the risk requirement.}
    \label{fig_openfig}
\end{figure}
Given a query image and a database, an image retrieval system 
selects the best matching candidates from the database. Image retrieval 
serves as the foundation for many computer vision applications, including face recognition \cite{cheng2019modified}, large-scale image classification \cite{warburg2021bayesian}, and person re-identification \cite{yao2019deep}. 

An image retrieval system typically 
works by first representing 
each query image as a vector in a high-dimensional space and then finding the nearest neighbors of the query image in a database. It is not surprising that building discriminative image representations is crucial, and for this reason massive research has been devoted to improving the representations of image features, spanning from early hand-crafted features to nowadays deep learning-based ones \cite{dubey2022a}. Nevertheless, like other deep learning-based applications, current image retrieval systems are black-box and data-driven, which renders them potentially unreliable due to limited training data and high model complexity. 

Meanwhile, reliability is essential for safety-critical applications ranging from autonomous driving to medical diagnosis, where it is vital to be aware of potential risks associated with the predictions. In the field of medical diagnosis, for instance, the user needs more than just a list of the best-matching candidates from disease databases. Instead, what they prefer or require is a cluster of retrieval results (or a retrieval set) with a specific level of risk, before concluding a diagnosis. The risk is defined as the probability of the retrieval set missing all true nearest neighbors of the query sample. See \cref{sec_risk_notion} for a formal definition. Conventional image retrieval systems are inadequate for these tasks as they do not provide any measure for risk control.

Uncertainty estimation has emerged as a risk measure for image retrieval systems. 
It involves estimating the level of uncertainty for both query and database samples. Existing approaches such as  \cite{warburg2021bayesian,chang2020data,zhang2021relative} use high uncertainty as an (informal) signal to indicate the potentially inaccurate predictions. 
However, there are limitations.
\emph{First}, the estimated uncertainty is merely a heuristic measure \cite{angelopoulos2022image} rather than a guarantee. This leaves end users with only an \textit{approximate} understanding about the prediction's risk. \emph{Second}, the estimated uncertainties do not consider the size of the retrieval set. That is, even if the retrieval set includes more candidates, which obviously increases its chance of containing the correct answer, the estimated uncertainty remains unaffected.

The above limitations motivate this work where we propose a novel image retrieval system called Risk Controlled Image Retrieval (\sysname).
\cref{fig_openfig} highlights the advantages of \sysname\ over the 
conventional uncertainty-aware image retrieval systems, by adhering to a pre-specified risk requirement with provable guarantee for the uncertainty measure. 
We consider a risk requirement that 
involves two factors: a risk level $\alpha$ and an error rate $\delta$, and require that, the probability that the assertion -- the risk of the retrieval set not including at least one true nearest neighbor is less than or equal to $\alpha$ -- occurs is greater than or equal to $1-\delta$. 
%
Specifically, given such a risk requirement, \sysname\ can not only adapt the size of the retrieval set, but also ensure that the retrieval set satisfies the requirement, offering a provable statistical guarantee to the underlying image retrieval system. 

Technnically, \sysname\ employs two modules, a Retrieval Set Size Adapter (\modulea) and a Risk Controller (\moduleb). 
The \moduleb\ calculates a scale $\kappa$ offline, which is then used to guide the \modulea\ to adjust the size of a retrieval set during inference. This collaboration enables the generation of a risk-controlled retrieval set.

We summarize the contributions of this paper as follows: 
\begin{enumerate}
    \item We for the first time utilize the common heuristic uncertainty estimation methods in a statistical manner to dynamically adjust the retrieval set size.
    \item We for the first time allow an image retrieval system to generate retrieval sets that meet a user-specified risk requirement.
    \item We demonstrate the effectiveness of \sysname\ through experimental results on four datasets: Stanford CAR-196, CUB-200, Pittsburgh, and ChestX-Det.
    \item The source code of \sysname\ is available at \url{https://github.com/ramdrop/rcir}.
\end{enumerate}

\section{Related Work}
\subsection{Image Retrieval}
Image retrieval often involves building a tagged database offline and searching it online, where images are represented by feature vectors. Therefore, image retrieval's effectiveness relies on the feature extractor's representation power. 
Initially, hand-crafted image features such as SIFT \cite{lowe1999object} were primarily used. However, with the superior performance of deep learning-based features from pretrained CNNs \cite{babenko2014neural}, hand-crafted features have become 
obsolete. Recent research on image retrieval demonstrates that deep learning-based features can be effectively learned end-to-end with ranking loss functions \cite{hadsell2006dimensionality,schroff2015facenet,sohn2016improved}. In this sense, image retrieval has evolved into a metric learning problem, with the aim of learning a mapping function that creates an embedding space where similar objects are positioned close together while dissimilar objects
farther apart. Loss functions, including contrastive loss \cite{hadsell2006dimensionality}, triplet loss \cite{schroff2015facenet}, and N-pairs loss \cite{sohn2016improved}, have been studied to enhance metric learning, which in turn benefits image retrieval.

\subsection{Uncertainty Estimation}

Deep learning has achieved tremendous success in numerous computer vision tasks, but its black-box working mechanism has raised concerns about its reliability. Uncertainty estimation is a way to quantify the confidence of the model in its prediction. Uncertainty of predictions can arise from either the data's inherent uncertainty or the model's uncertainty, known as aleatoric and epistemic uncertainty, respectively \cite{kendall2017uncertainties}. To quantify epistemic uncertainty, variational inference based methods such as Monte Carlo Dropout are developed to approximate Bayesian Neural Networks, where the weights of the networks are modeled as distributions. Ensemble method \cite{fort2019deep} initiates multiple instances of the same model and then takes the variance of predictions as an uncertainty level. On the other hand, researchers typically estimate aleatoric uncertainty by learning it via an additional head that is parallel to the network.

Specific to the field of image retrieval, DUL \cite{chang2020data} learns aleatoric uncertainty by constructing a stochastics embedding space where the uncertainty is regularized by a KL divergence loss. BTL \cite{warburg2021bayesian} utilizes a Bayesian loss function that enforces triplet constraints on stochastic embeddings. \cite{taha2019exploring} shows MCD can be employed in image retrieval to quantify epistemic uncertainty. Current research on uncertainty estimation in image retrieval is primarily concerned with quantifying the likelihood of a pair of images being similar or dissimilar, albeit with only a heuristic estimate of uncertainty \cite{angelopoulos2022image}. Our objective is to retrieve a set of images guaranteed to contain the true nearest neighbors of a given query sample with a user-specified risk requirement.

\subsection{Conformal Prediction}

The framework of conformal prediction (CP) provides a  guarantee to the correctness of a prediction \cite{vovk2005algorithmic}. Given a set of data points $(X_i,Y_i)\in \mathcal{Q}\times \mathcal{Y}$, for $i=1,..,n$, on which a classifier $f$ is trained, instead of returning a response $f(X_{n+1})$, the split conformal prediction, a popular CP method, returns a prediction set $\mathcal{C}(X_{n+1})\subseteq \mathcal{Y}$ such that $P(Y_{n+1}\in \mathcal{C}(X_{n+1}))\geq 1-\alpha$ for $\alpha\in [0,1]$ an user-specified error level. It defines a score function $s$, computes a quantile $t$ for  $X_{n+1}$ over a set $\mathcal{D}_c$ of calibration dataset, and then use $t$ as a threshold to determine $\mathcal{C}(X_{n+1})$. The statistical guarantee depends on the exchangability (a concept that is slightly relaxted from the i.i.d. assumption) in training and calibration datasets  and query sample, as well as the hypothesis testing. 
 
Beyond the 
initial study on classfication, conformal quantile regression \cite{romano2019conformalized} enhances CP to deliver prediction intervals with guaranteed accuracy. Building on this idea, DFRP \cite{bates2021distribution} generates risk-controlled prediction set for image classification, and DFUQ \cite{angelopoulos2022image} proposes generating risk-guaranteed intervals in image regression. 
Different from these works on image classification or regression models, where the models are trained with one-to-one labels, our work focuses on image retrieval, where the model is trained using pairwise relations. 
Furthermore, the uncertainty estimation methods for regression tasks are distinct from those used in image retrieval since true labels are not available \cite{chang2020data}. 
These differences necessitate a thorough revisit to the formulation of the risk, as well as the formal proof on the guarantee, within this new paradigm. 


\section{Method}

\begin{figure}[t]
    \begin{center}
       \includegraphics[width=\linewidth]{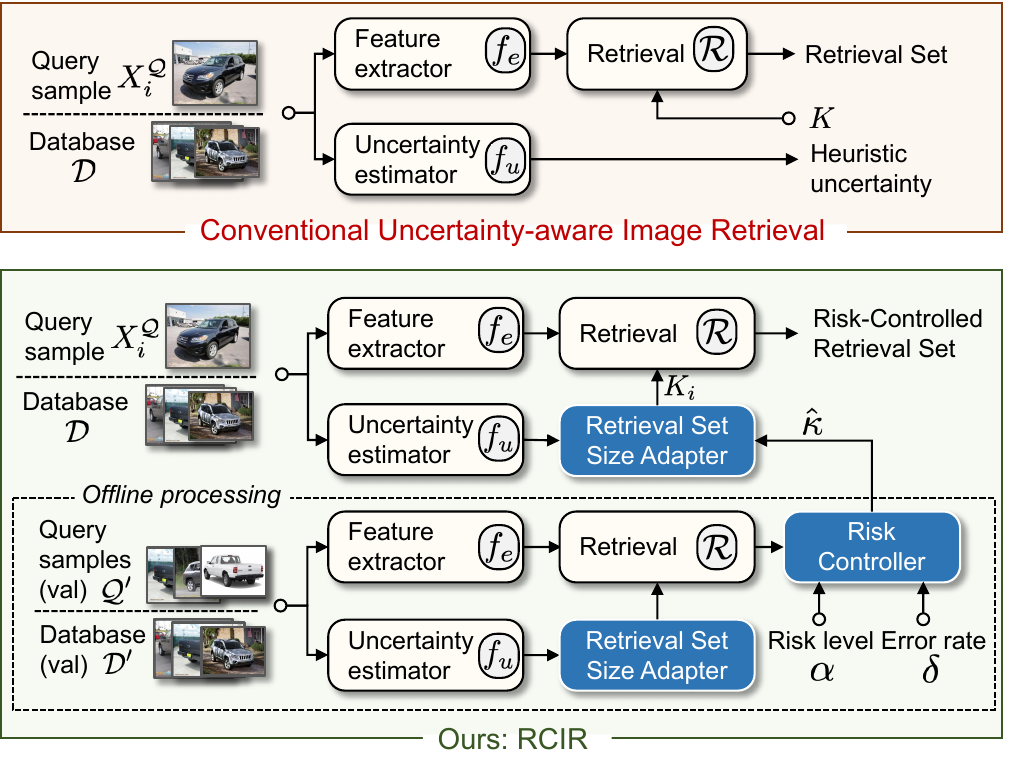}
    \end{center}
       \caption{Diagrams of a conventional image retrieval system and our \sysname: \textcolor{color_02}{Blue blocks}\raisebox{-0.11cm}{\includegraphics[width=14pt]{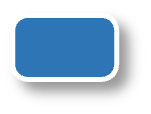}} (i.e., Retrieval Set Size Adapter and Risk Controller) highlight our contributions. Please see the text below for a detailed explanation.} 
       \label{fig_diagram}
 \end{figure}

\cref{fig_diagram} presents the workflows of a conventional uncertainty-aware image retrieval system \cite{warburg2021bayesian}, as well as our \sysname. The conventional uncertainty-aware image retrieval system first uses 
a
feature extractor $f_e$ to project 
the query and database samples into embeddings, and then applies
a
retrieval function $\mathcal{R}$ to obtain the retrieval set, which forms of the $K$ nearest neighbors of the query sample in the embedding space. Based on this, the uncertainty estimator $f_u$ 
computes a heuristic uncertainty for each query sample. Our \sysname\ differs from the conventional one in that,  the retrieval set is \emph{risk-controlled}: during offline processing, given the predefined risk requirements, i.e., the risk level $\alpha$ and the error rate $\delta$, we first use a \moduleb\ to evaluate the risk of $\mathcal{R}$ under various  hypothesis on a scale $\kappa$ in order to determine the best scale 
$\hat{\kappa}$. Then, during inference, given the $i^{th}$  query sample $X_{i}$, an \modulea\ generates the retrieval set of size $K_i$, accroding to $\hat{\kappa}$, such that 
\begin{equation}\label{eq:main}
    P(\rho(\mathcal{R}) \leq \alpha) \geq 1-\delta
\end{equation}
where $\rho(\mathcal{R})$ is the risk of the retrieval set $\mathcal{R}(X)$ for a query set $\mathcal{Q}$ where $X\in \mathcal{Q}$. The definitions of $\mathcal{R}(X)$ and $\rho(\mathcal{R})$ will be elicited below in Eqs. (\ref{eq_rk}) and (\ref{eq_risk}), respectively. 

In the following, we first describe the conventional image retrieval pipeline, followed by the uncertainty-aware variant before moving into our risk controlled enhancement. 


\subsection{Image Retrieval}
In image retrieval, a feature extractor $f_e$ is trained to 
project high-dimensional images onto a lower-dimensional embedding space. In the embedding space, similar samples are positioned closely together, whereas dissimilar samples are positioned far apart. Assume that there is a dataset $\{\mathcal{Q}, \mathcal{D}\}$, where $\mathcal{Q}$ denotes the query set and $\mathcal{D}$ the database. Given the $i^{th}$  query sample $X_{i}^\mathcal{Q} \in \mathcal{Q}$ (superscript denotes which set the sample belongs to), its most similar samples $\{Y_{i, j} | j=1,2,.., K\}$ are retrieved from $\mathcal{D}$ as 
\begin{equation}
    \label{eq_r_conventional}
   \{Y_{i, j} | j=1,2,..,K\} = \mathcal{R}_{[K, f_{e}]} (X_{i}^\mathcal{Q}),
\end{equation}
where $\mathcal{R}_{[K, f_{e}]}$ represents a retrieval function conditioned on $K$ and $f_{e}$:
\begin{equation}
    \label{eq_rk}
\begin{aligned}
    \mathcal{R}_{[K, f_{e}]} (X_i^\mathcal{Q}) = &
     \{ x \vert x \in \mathcal{P}, \mathcal{P} \subseteq \mathcal{D}, \vert \mathcal{P} \vert= K, \\ & d\left[f_e (X_i^{\mathcal{Q}}), f_e (x)\right] \leq  d\left[f_e (X_i^{\mathcal{Q}}), f_e (z)\right],\\ & \forall z \in \mathcal{D} \backslash  \mathcal{P} \},
\end{aligned}
\end{equation}
where $d$ denotes a distance metric function, and $K$ denotes the number of the retrieved candidates,  $f_e$ denotes the feature extractor.  The above image retrieval system has been the \textit{de facto} standard in the image retrieval community. However, it only provides a list of best-matching candidates without indicating their reliability. The absence of reliability information can be problematic in situations where failures may have high-stakes consequences. 

\subsection{Uncertainty-aware Image Retrieval}
\label{sec_uncertainty}
Uncertainty estimation techniques mitigate this issue by providing a heuristic uncertainty for each query and database sample. Existing uncertainty-aware image retrieval systems include three methodologies: 1) estimate model uncertainty by BNNs \cite{gal2016dropout}, 2) estimate uncertainty by an ensemble of models, and 3) predict uncertainty by a deterministic model \cite{warburg2021bayesian}. Since \sysname\ utilizes heuristic uncertainty, we briefly introduce them.
\subsubsection*{Uncertainty Estimated by a BNN}
Bayesian approaches treat the weights of a neural network as distributions instead of deterministic values. Since obtaining the analytical posterior distribution for the weights is intractable due to the difficulty of acquiring evidence, researchers have utilized various methods to address this challenge, with variational inference being the most widely employed method, e.g., Monte Carlo Dropout (MCD) \cite{gal2016dropout}. MCD assumes a mixed Gaussian distribution on each weight, which makes the sampling of weights equivalent to applying dropout operations. In our image retrieval setting, we apply dropout to all conventional layers of the feature extractor with a dropout rate $p$. Feature $\mu$  and the heuristic uncertainty $\sigma^2$ are obtained by applying dropout at test time and $T$ forward propagations:
\begin{equation}
    \label{eq_uncertainty_bnn}
    \begin{aligned}
    \mu = \frac{1}{T} \sum_{t=1}^{T} f_{e_t \sim \theta}(X),
    \sigma^2 = \frac{1}{T} \sum_{t=1}^{T} (f_{e_t \sim \theta}(X)-\mu)^2.\\
    \end{aligned}
\end{equation}
where $\theta$ denotes the weight posterior distribution.

\subsubsection*{Uncertainty Estimated by an Ensemble} Ensemble method \cite{fort2019deep} trains multiple instances of a same deterministic model, each with random initial weights. During inference, predictions of all instances are averaged, and the uncertainty is obtained as the variance of the predictions. Provided that there are $N$ instances, the mean and variance of the predictions are:
\begin{equation}
    \label{eq_uncertainty_ensemble}
    \begin{aligned}
    \mu = \frac{1}{N} \sum_{i=1}^{N} f_{e_i}(X),
    \sigma^2=\frac{1}{N} \sum_{i=1}^{N} (f_{e_i}(X)-\mu)^2.\\
    \end{aligned}
\end{equation}

\subsubsection*{Uncertainty Estimated by a Single Deterministic Model}
Multiple feed-forward propagations in MCD can cause overhead, and using multiple model instances can result in increased memory usage. In comparison, a single deterministic model is an attractive approach for estimating uncertainty. \cite{warburg2021bayesian} constructs a Gaussian distribution for each feature, i.e., $f_e(X) \sim \mathcal{N}(\mu, \sigma^2)$. A Bayesian triplet loss is introduced to enforce contrastive learning among these probabilistic features. The network is built upon a common image retrieval model by adding a variance head $f_u$  parallel to the mean head $f_e$. Once trained, the model can output $\mu$ and $\sigma^2$ with one forward propagation:
\begin{equation}
    \label{eq_uncertainty_sdm}
    [\mu, \sigma^2] = [f_{e}(X), f_{u}(X)].
\end{equation}

In summary, uncertainties of features are obtained by a BNN-based method, an ensemble or a single deterministic model. 
Nevertheless, the existing uncertainty-aware image retrieval methods are problematic in two senses:
\begin{enumerate}
    \item They only provide a \textit{heuristic} measure of uncertainty \cite{angelopoulos2022image} rather than a guarantee. That said, even with the estimated uncertainty, the retrieval set \textit{cannot} be interpreted with a likelihood of encompassing the true nearest neighbors of a given query sample.
    \item The retrieval set's uncertainty is represented by the query sample's uncertainty \cite{warburg2021bayesian}, which does not consider the retrieval set size. As a result, the same uncertainty value is assigned to a retrieval set with  one sample as to a retrieval set with ten samples. However, a larger retrieval set has a higher likelihood of capturing the true nearest neighbors compared to a smaller one.
\end{enumerate}

We propose \sysname\ to overcome these limitations: \sysname\ takes as input the user-specified risk level $\alpha$ and error rate $\delta$, and outputs a risk-controlled retrieval set guaranteed to contain at least one true nearest neighbor of the query sample with a probability $1-\alpha$ and an error rate $\delta$. We describe \sysname\ in what follows.

\subsection{Risk Controlled Image Retrieval}
\subsubsection{Problem Setting}
\label{sec_risk_notion}
We denote the risk of an image retrieval system $\mathcal{R}$ by $\rho(\mathcal{R})$, defined as the expected value, over a set $\mathcal{Q}$ of query samples, on the case that the retrieval set misses \emph{all} true nearest neighbors:
\begin{equation}
    \label{eq_risk}
    \rho(\mathcal{R}) = \mathds{E}_{X \in \mathcal{Q}}[\ell(\mathcal{R}(X), \mathcal{S}(X))],
\end{equation}
where $\mathcal{R}$ denotes a generic retrieval function as in \cref{eq_rk}, $\mathcal{S}(X)$ means retrieving all true nearest neighbors of $X$, and $\ell$ indicates if the retrieval set misses all true nearest neighbors:
\begin{equation}
\label{eq_loss}
\ell(\mathcal{R}(X), \mathcal{S}(X)) = \mathds{1}(\mathcal{R}(X) \cap \mathcal{S}(X) =  \emptyset).
\end{equation}

The risk function $\rho(\cdot)$  evaluates the performance of a retrieval system $\mathcal{R}$ on a query set $\mathcal{Q}$, and is bounded between $0$ and $1$: $\rho(\mathcal{R})=0$ occurs when the retrieval set $\mathcal{R}(X)$ covers at least one true nearest neighbor for all $X$ in $\mathcal{Q}$, and $\rho(\mathcal{R})=1$ happens when the retrieval set $\mathcal{R}(X)$ never covers any true nearest neighbor for any $X$ in $\mathcal{Q}$. 

Based on the above definition, the goal of our \sysname\ is to show that the risk $\rho(\mathcal{R})$ can be provably guaranteed to be below a predefined level $\alpha$ with a probability of $1-\delta$, in the form of Eq. (\ref{eq:main}). In \sysname, this is achieved by employing two components: an \modulea\ and a \moduleb, where the \moduleb\  determines a scale $\hat{\kappa}$ (Eq. (\ref{eq_k_hat})), whereas the \modulea\ uses the scale to adaptively determine the retrieval set (Eq. (\ref{eq_pi})). The guarantee is established formally in Theorem~\ref{thm_ucb} (Eq. (\ref{eq_guarantee})). 


\subsubsection{Retrieval Set Size Adapter}
In practice,  without 
compromising accuracy, end users 
prefer smaller retrieval sets for `easy' query samples, compared to `hard' ones, in order to save post-processing time. To achieve this, we propose adapting the retrieval set size based on the query's \textit{difficulty level}. While appealing, achieving this goal is not straightforward and has been barely explored in existing literature.

The first question we need to answer is: \textit{what is the appropriate metric to measure a query's difficulty level?} Our investigation shows that although existing heuristic uncertainties do not provide guarantees, they 
correlate with recall@1 (see \cref{fig_multiple_ece}). This discovery leads us to hypothesize that common heuristic uncertainty can serve as a coarse measure of a query's difficulty level.

Moving on to the second question: \textit{If such a metric exists, what would be the principled way of building the relation between this metric and the retrieval set size?} For this, we have identified a simple yet effective solution: 
\begin{equation}
    \label{eq_pi}
    K_i = \left\lceil \kappa \cdot \Phi[f_u{(X_i^\mathcal{Q})]} \right\rceil,
\end{equation}
where $\kappa \in \mathds{R}^+$  denotes a scale, which will be computed by the \moduleb\ (details to explained below, with the best $\kappa$ as $\hat{\kappa}$),  $f_u$ is a heuristic uncertainty estimator, $\Phi$ 
normalises the result into $[0, 1]$, and $\left\lceil \cdot \right\rceil $ 
rounds up to the nearest integer.

We denote the mapping function, \cref{eq_pi}, as the \modulea, which adjusts the retrieval set size based on the uncertainty of each query sample and a scale $\kappa$. Then, an image retrieval function with the \modulea\ is  conditioned on $\kappa$, $f_u$ and $f_e$:
\begin{equation}
    \label{eq_r_adaptive}
    \{Y_{i, j} | j=1,2,..,K_i\} = \mathcal{R}_{[\kappa, f_{u}, f_{e}]} (X_{i}^\mathcal{Q}),
\end{equation}
and its risk is denoted as  $\rho(\mathcal{R}_{[\kappa, f_{u}, f_{e}]})$. Since $f_u$ and $f_e$ are fixed once trained, $\rho(\mathcal{R}_{[\kappa, f_{u}, f_{e}]})$ only changes with respect to $\kappa$. Thereafter, we may also refer to the risk as $\rho(\kappa)$.

\subsubsection{Risk Controller}

The \modulea\ alone is insufficient to provide 
provable
guarantee on risk freeness for an image retrieval system. To complement it, we introduce \moduleb, which enhances 
the \modulea\ with risk awareness.

Given the given risk level is $\alpha$ and the error rate is $\delta$, the \moduleb\ is responsbile for determining the scale $\kappa$ such that the risk of the retrieval system is below $\alpha$ with  probability greater than $1-\delta$. The general idea is as follows. Because $\rho(\kappa)$ is unknown, we consider instead its upper confidence bound (UCB) $\hat{\rho}^+(\kappa)$, which can be determined by applying Hoeffding's inequality (Eq.~\ref{equ:hoeffding}). Because the UCB $\hat{\rho}^+(\kappa)$ is greater than the risk $\rho(\kappa)$ with high probability (Eq. (\ref{eq_ucb})), and the fact that the risk function is monotone nonincreasing (Lemma~\ref{lemma_monotone}), we can prove in Theorem~\ref{thm_ucb} that, as long as the scale $\hat{\kappa}$ is the smallest one such that $\hat{\rho}^+(\kappa) \leq \alpha$ for all $\kappa>\hat{\kappa}$, the guarantee holds on $\hat{\kappa}$ (Eq. (\ref{eq_guarantee})). It is not hard to see that Eq. (\ref{eq_guarantee}) is an instantiation of Eq. (\ref{eq:main}). 

\begin{lemma}
    \label{lemma_monotone}
    Given a dataset $\{\mathcal{Q}, \mathcal{D}\}$, with a fixed feature extractor $f_e$ and a fixed uncertainty estimator $f_u$, the risk function $\rho(\mathcal{R}_{[\kappa, f_{u}, f_{e}]})$ is a monotone nonincreasing function with respect to the scale $ \kappa \in \mathds{R}^+ $. 
\end{lemma}

\begin{proof}[Proof]
    For any $i^{th}$ query $X_i^\mathcal{Q}$, if $\kappa_1 > \kappa_2$, then $K_{i, \kappa_1} \geq K_{i, \kappa_2}$  by the mapping function Eq. (\ref{eq_pi}) as $f_u$ is fixed. This means $\vert \mathcal{R}_{[\kappa_1, f_{u}, f_{e}]}(X_i^\mathcal{Q})\vert \geq \vert \mathcal{R}_{[\kappa_2, f_{u}, f_{e}]}(X_i^\mathcal{Q})\vert$, accroding to Eq. (2). Note that,  from Eq. (\ref{eq_rk}),  the risk function is monotonic with respect to $K$, the size of the retrieval set. 
    Then, because in a retrieval system, the candidates are retrieved based on a consistent distance metric $d$ (see Eq. (\ref{eq_rk})), we have $$\mathcal{R}_{[\kappa_2, f_{u}, f_{e}]}(X_i^\mathcal{Q}) \subseteq  \mathcal{R}_{[\kappa_1, f_{u}, f_{e}]}(X_i^\mathcal{Q}).$$ 
    One step further, according to the loss definition in Eq. (\ref{eq_loss}), $$\ell(\mathcal{R}_{[\kappa_2, f_{u}, f_{e}]}(X_i^\mathcal{Q}), \mathcal{S}(X_i^\mathcal{Q})) \geq \ell(\mathcal{R}_{[\kappa_1, f_{u}, f_{e}]}(X_i^\mathcal{Q}), \mathcal{S}(X_i^\mathcal{Q})).$$
    Finally, by the risk definition in Eq. (\ref{eq_risk}), we have 
    $$\rho(\mathcal{R}_{[\kappa_2, f_{u}, f_{e}]}) \geq \rho(\mathcal{R}_{[\kappa_1, f_{u}, f_{e}]}),$$
    which concludes the proof. 
    \end{proof}

\begin{theorem}
\label{thm_ucb}
    Assume the upper confidence bound of the risk, which we denote by $\hat{\rho}^+(\kappa)$,  is accessible, and  $\forall \kappa \in \mathds{R}^+ $, 
    \begin{equation}
        \label{eq_ucb}
        P(\rho(\kappa) \leq \hat{\rho}^+(\kappa)) \geq 1-\delta.
    \end{equation}
    Let $\hat{\kappa}$  denotes the smallest $\kappa$ such that for any $\kappa > \hat{\kappa}$ we have $\hat{\rho}^+(\kappa) \leq \alpha$, i.e.,
    \begin{equation}
        \label{eq_k_hat}
        \hat{\kappa} = \inf\{\kappa: \hat{\rho}^+(\kappa^\prime) < \alpha, \forall \kappa^\prime > \kappa\},
    \end{equation}
    Then, we have that 
    \begin{equation}
    \label{eq_guarantee}
    P(\rho(\hat{\kappa}) \leq \alpha) \geq 1-\delta. 
    \end{equation}
\end{theorem}

\begin{figure}[tp]
    \centering
    \includegraphics[width=0.5\columnwidth]{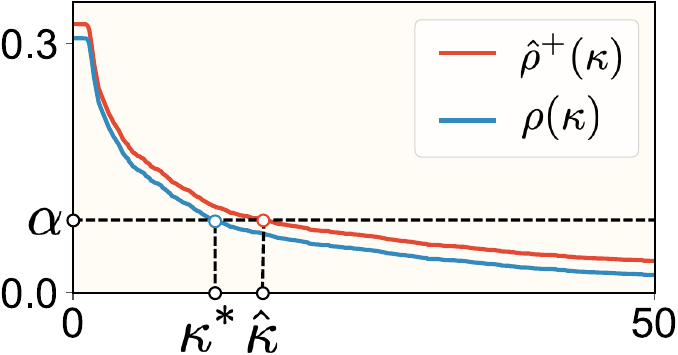} 
    \caption{The $\rho-\kappa$ curve on the CAR-196 calibration set: the risk  $\rho(\kappa)$ is monotone nonincreasing with $\kappa$.}
    \label{fig_loss_curve}
\end{figure}

\begin{proof}[Proof]
    Let $\kappa^*$ be the smallest value of $\kappa$ that ensures the risk is less than $\alpha$, as given by the following equation: 
    \begin{equation}
        \label{eq_k_star}
        \kappa^* = \inf\{\kappa: \rho(\kappa) \leq \alpha\},
    \end{equation}
    Then, due to the continuity and nonincreasing (Lemma~\ref{lemma_monotone}) of $\rho(\kappa)$, it follows that $\rho(\kappa^*)=\alpha$. 
    To make it easier to understand, we visualize in \cref{fig_loss_curve} the risks $\rho(\kappa^*)$, $\rho(\hat{\kappa})$ and $\alpha$ using the CAR-196 calibration set. Although it is an specific example, it does not lose the generality. 
    
    
    We define two events: the event $A$, which occurs when $\rho (\hat{\kappa}) > \alpha$, and the event $B$, which occurs when $\rho(\kappa^*) > \hat{\rho}^+(\kappa^*)$. We show that $A$ is a sufficient condition for $B$. From $\rho (\hat{\kappa}) > \alpha$, we have that  $\hat{\kappa} < \kappa^*$ (due to the monotonicity of $\rho(\kappa)$), which implies $ \hat{\rho}^+(\kappa^*) < \alpha $ (according to \cref{eq_k_hat}). Now, because  $\rho(\kappa^*)=\alpha$ and $\hat{\rho}^+(\kappa^*)<\alpha$, we have that $\rho(\kappa^*) > \hat{\rho}^+(\kappa^*)$.  
    
    By the fact that $A$ is a sufficient condition for $B$, we have that  $P(A) \leq P(B)$. According to \cref{eq_ucb}, we have $P(B)<\delta$. The combination of $P(A) \leq P(B)$ and $P(B)<\delta$ leads to  $P(A) < \delta$. Consequently, $P(\overline{A}) \geq 1- \delta$, i.e., $P(\rho(\hat{\kappa}) \leq \alpha) \geq 1-\delta$.
\end{proof}

\begin{algorithm}[t]
    \caption{ Compute ${\hat{\kappa}}$.}
    \label{alg_regress_k}
    \textbf{Input}: {Calibration set $\{ \mathcal{Q}^\prime, \mathcal{D}^\prime\}$, risk level $\alpha$, error rate $\delta$, trained feature extractor $f_{e}$, trained heuristic uncertainty estimator $f_{u}$
    }, step $\Delta \kappa$ \\
    \textbf{Output}: ${\hat{\kappa}}$
    \begin{algorithmic}[1] 
    \STATE $\kappa \leftarrow 1$,  $\hat{\rho}^+(\kappa)  \leftarrow$ 1
    \WHILE{ $ \hat{\rho}^+(\kappa)  > \alpha$}
    \STATE $\kappa = \kappa + \Delta \kappa $\\
    \FOR{$i$ in $1,2,..,\vert \mathcal{Q^\prime} \vert$} 
    \STATE $\tilde{K_i} = \left\lceil \kappa \cdot \Phi[f_u{(X_i^\mathcal{Q^\prime})]} \right\rceil$
    \STATE $ \ell_{i}=  \mathds{1}\left[ \mathcal{R}_{[\tilde{K_i}, f_{e}]} (X_{i}^\mathcal{Q^\prime})\cap \mathcal{S}(X_i^\mathcal{Q^\prime}) =  \emptyset \right] $ 
    \ENDFOR
    \STATE $\hat{\rho}(\kappa) =  \frac{1}{\vert \mathcal{Q^\prime} \vert}\sum_i^{\vert \mathcal{Q^\prime} \vert}\ell_{i}$ 
    \STATE {$\hat{\rho}^+(\kappa)  = \hat{\rho}(\kappa) + \sqrt{\frac{1}{2 n} \log \frac{1}{\delta}}$  }
    \ENDWHILE
    \STATE  $\hat{\kappa} = \kappa - \Delta \kappa $
    \STATE \textbf{return} $\hat{\kappa}$
    \end{algorithmic}
\end{algorithm}

With \cref{eq_k_hat}, we can proceed to determine an upper confidence bound $\hat{\rho}^+(\kappa)$. According to \cite{bates2021distribution}, Hoeffding’s inequality applies to our risk function $\rho(\kappa)$, which is bounded by $1$. We denote the risk on the calibration set by $\hat{\rho}(\kappa)$. Hoeffding’s inequality indicates 
\begin{equation}
P(\hat{\rho}(\kappa) - \rho(\kappa) \leq -x) \leq e^{-2nx^2},
\end{equation}
which implies an upper confidence bound 
\begin{equation}\label{equ:hoeffding}
    \hat{\rho}^{+}(\kappa)=\hat{\rho}(\kappa)+\sqrt{\frac{1}{2 n} \log \left(\frac{1}{\delta}\right)}.
\end{equation}

The \moduleb\ calculates $\kappa$ by following the steps outlined by \cref{alg_regress_k}: Given a calibration set, $\alpha$, $\delta$, $f_{e}$, $f_{u}$ and $\Delta \kappa$,  the \moduleb\ computes $\hat{\kappa}$ by iteratively increasing $\kappa$ until the upper confidence $\hat{\rho}^+(\kappa)$ bound is below $\alpha$. 

With all key steps explained, we summarize the pipeline of \sysname\ as follows: Given a predefined risk requirement, i.e., risk level $\alpha$ and error rate $\delta$, the \moduleb\ uses \cref{alg_regress_k} to compute ${\hat{\kappa}}$. Once ${\hat{\kappa}}$ is obtained, the \modulea\ generates the retrieval set size $K_i$, and the image retrieval function retrieves the top $K_i$ candidates. The set of these candidates is the risk-controlled retrieval set as it is ensured to have at least one true nearest neighbor with a probability of $1-\alpha$, and that the chance of the assert fail is limited to $\delta$.

\section{Experiments}

\subsection{Datasets}
\noindent \textbf{CUB-200} \cite{wah2011caltech} contains 11,788 images of 200 classes. Each class has at least 50 images. Following \cite{warburg2021bayesian}, we use the first 100 classes as a training set and the other 100 as a test set. 

\noindent \textbf{Stanford CAR-196} \cite{krause20133d} contains 16,185 images of 196 classes. Each class has at least 80 images. Following \cite{warburg2021bayesian}\cite{musgrave2020metric}, we use the first 98 classes as the training set and the other 98 classes as the test set. 

\noindent \textbf{Pittsburgh} \cite{torii2013visual} is a large image database from Google Street View. Following the split of NetVLAD \cite{arandjelovic2016netvlad}, we adopt the subset that consists of 10k samples in the training, validation and test split. 

\noindent \textbf{ChestX-Det} \cite{lian2021structure} is a subset of the public dataset NIH ChestX-ray14, and contains 3543 images with 14 classes.
We choose samples of the six classes as the training set, and the rest as the test set. 

For all datasets, we follow \cite{angelopoulos2022imagetoimage} and randomly select 50\% images from the test set to form the calibration set.

\subsection{Implementations}
\sysname\ is versatile and can be applied to any image retrieval system that provides heuristic uncertainty. To show the effectiveness of \sysname, we first build three conventional image retrieval systems,  \textbf{BTL},  \textbf{MCD} and \textbf{Deep Ensemble}, based on the model used in \cite{warburg2021bayesian}. Each system has a feature extractor and an uncertainty estimator as described in \cref{sec_uncertainty}. BTL, MCD and Deep Ensemble retrieve fixed $K$ database samples when given a query. For comparing purposes, we also include a \textbf{Deterministic} image retrieval system that only has a feature extractor. 




All models are trained using the Adam optimizer with an initial learning rate of $10^{-5}$ and an exponential learning rate scheduler with gamma $0.99$. The weight decay is set to $10^{-3}$. The feature dimension is $2048$ and the variance dimension is $1$. We adopt a hard mining strategy, feeding models with triplets that violate the triplet margin \cite{warburg2021bayesian}. More information about the network and training can be found in the supplementary materials.

\begin{figure}[t]
    \centering
    \includegraphics[width=\columnwidth]{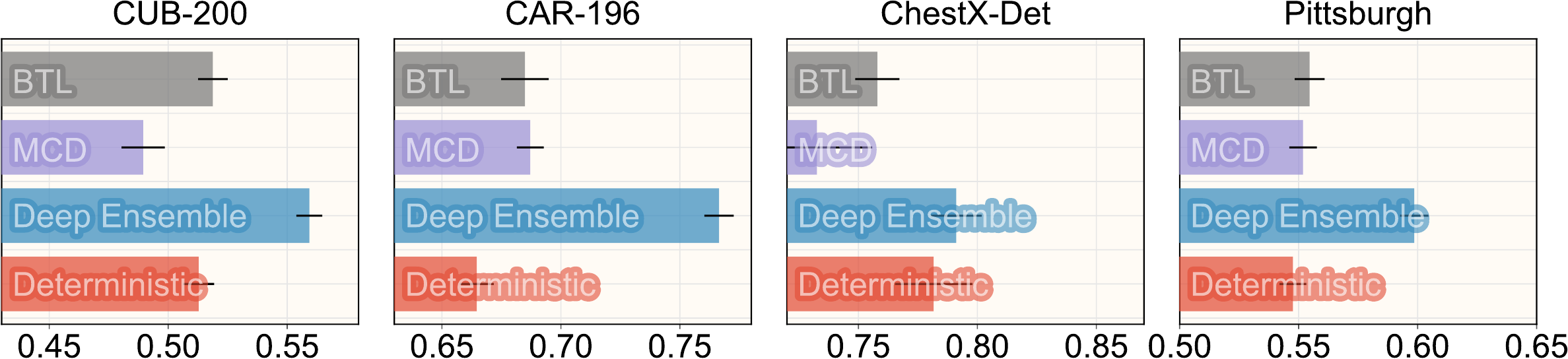} 
    \caption{The Recall@$1$ results of different methods on various test sets. The error bars represent the standard deviation of the results of $10$ trials.}
    \label{fig_recallk}
\end{figure}

\begin{figure}[t]
    \centering
    \includegraphics[width=\columnwidth]{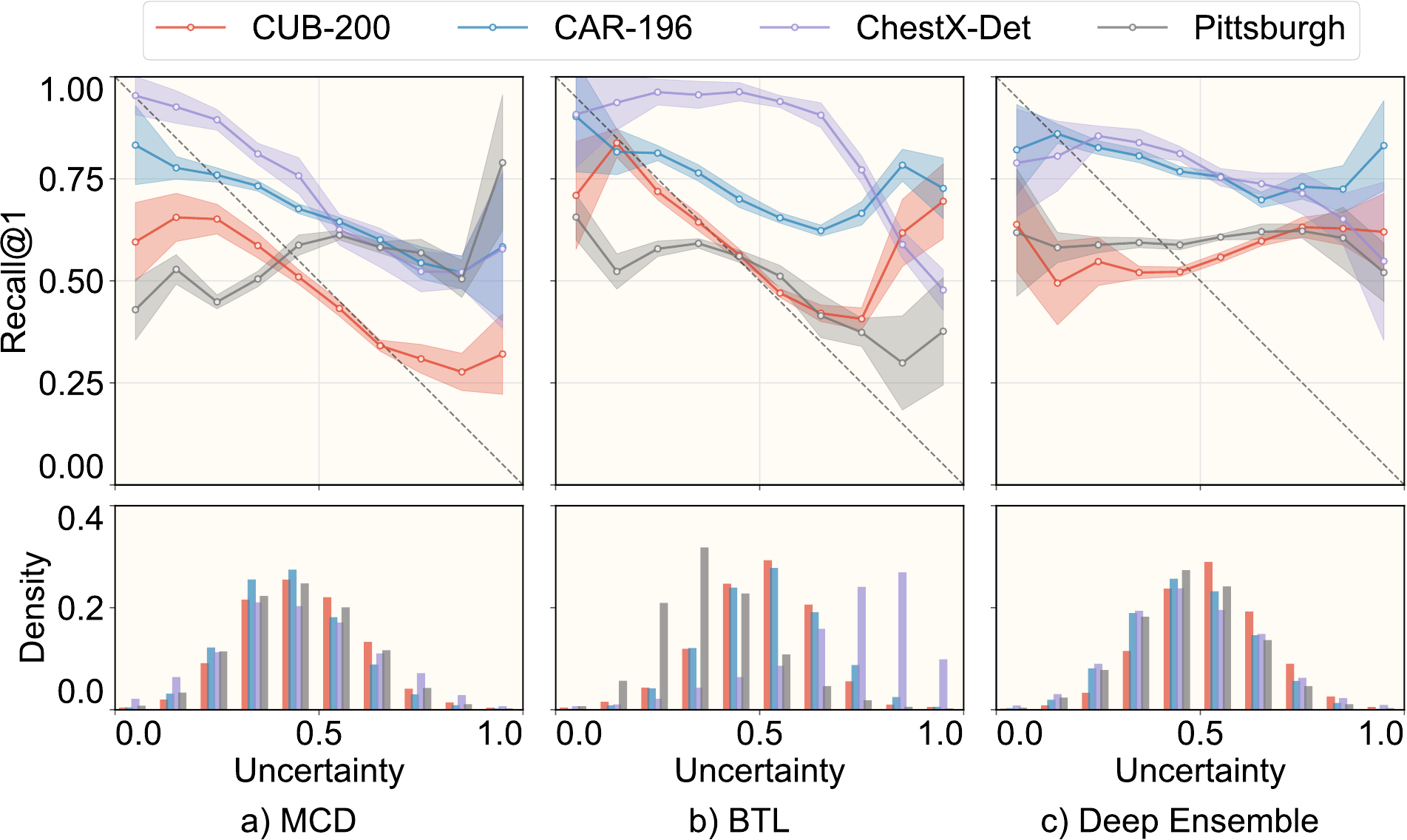} 
    \caption{The reliability diagrams on different test sets. The dashed line denotes the ideal calibration line. (The shadows represent the standard deviation of the results of $10$ trials, and the same goes for the other figures)}    
    \label{fig_multiple_ece}
\end{figure}

\subsection{Metrics}
We first follow the common evaluation steps \cite{warburg2021bayesian} for image retrieval systems, adopting \textbf{recall@1} to evaluate the retrieval performance of the feature extractor and \textbf{reliability diagram} \cite{guo2017on} to assess the uncertainty quality of the uncertainty estimator.
Then, we use \textbf{empirical risk}, which is the risk calculated on the test set, to evaluate the risk control performance.
In addition, \textbf{retrieval set size} is used to verify that \sysname\ does not resort to large retrieval sets to lower the empirical risk. The retrieval set size should be minimized while the risk is controlled. Because large retrieval sets tend to include more negative samples and are inefficient for downstream processing. We also provide \textbf{qualitative visualization} to show more insights into the image retrieval system. Please see the supplementary materials for more information on these metrics.

\subsection{Results}

\subsubsection{Recall@1 (Image Retrieval Performance)}
We begin by evaluating the retrieval performance of various methods. The Recall@$1$ of different image retrieval systems on various test sets are presented in \cref{fig_recallk}. BTL achieves a higher recall@1 than Deterministic on most datasets, suggesting that a feature extractor can benefit from uncertainty-aware training. Nevertheless, MCD's performance is inconsistent when compared to Deterministic. The inconsistency in performance may stem from the varying impact of dropout layers on the representation power, depending on the dataset's scale. Deep Ensemble achieves the best performance, which can be attributed to the fact that it utilizes multiple models to average out noises.

\subsubsection{Reliability Diagram (Uncertainty Estimation Error)} 
\cref{fig_multiple_ece} depicts the reliability diagrams of the uncertainty estimator of different image retrieval systems across datasets. It can be seen that none of these curves conform to the ideal calibration line, indicating that the heuristic uncertainty itself is not well-calibrated. Moreover, the gaps between the curves and the dashed line vary across datasets, implying that existing uncertainty estimation methods cannot consistently perform across different datasets.

\begin{figure}[h]
    \centering
    \includegraphics[width=1\columnwidth]{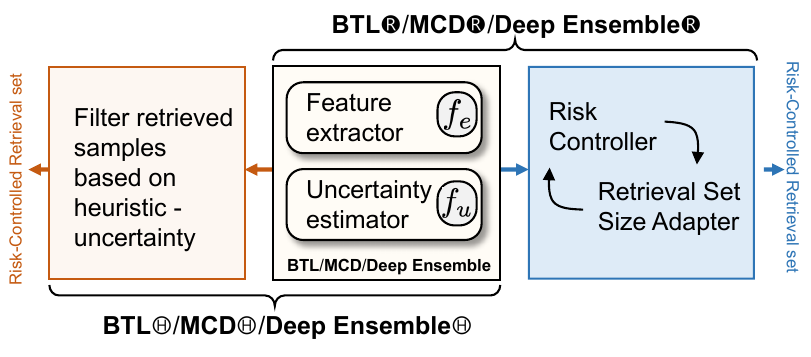} 
    \caption{The data flow of our methods, \textbf{BTL}\sysr\, \textbf{MCD}\sysr\ and \textbf{Deep Ensemble}\sysr, and the comparing methods, \textbf{BTL}\sysh, \textbf{MCD}\sysh\ and \textbf{Deep Ensemble}\sysh.}    
    \label{fig_methods_setting}
\end{figure}

Next, we build \sysname\ by plugging our \modulea\ and \moduleb\ modules into conventional image retrieval systems. We denote the three resulting \sysname\ systems as \textbf{BTL}\sysr\, \textbf{MCD}\sysr\ and \textbf{Deep Ensemble}\sysr. Meanwhile, to evaluate the practical usefulness of heuristic uncertainty, we normalize heuristic uncertainties to the range $[0, 1]$ based on their statistics on the calibration set. Then each query-candidate pair's uncertainty is calculated as the sum of their individual uncertainties \cite{warburg2021bayesian}. This will result in three comparing methods: \textbf{BTL}\sysh, \textbf{MCD}\sysh\ and \textbf{Deep Ensemble}\sysh. \cref{fig_methods_setting} depicts the data flow of the aforementioned methods.

For fairness, $\times$\raisebox{-0.08\height}{\sysh} will retrieve as many candidates as $\times$\raisebox{-0.08\height}{\sysr} can achieve, then only pairs with uncertainties below $1-\alpha$ will be kept. We adopt a step $\Delta \kappa=0.2$ for $\times$\raisebox{-0.08\height}{\sysr}.

\subsubsection{Empirical Risk (Risk Control Effectiveness)} 

\begin{figure}[t]
    \centering
    \includegraphics[width=\columnwidth]{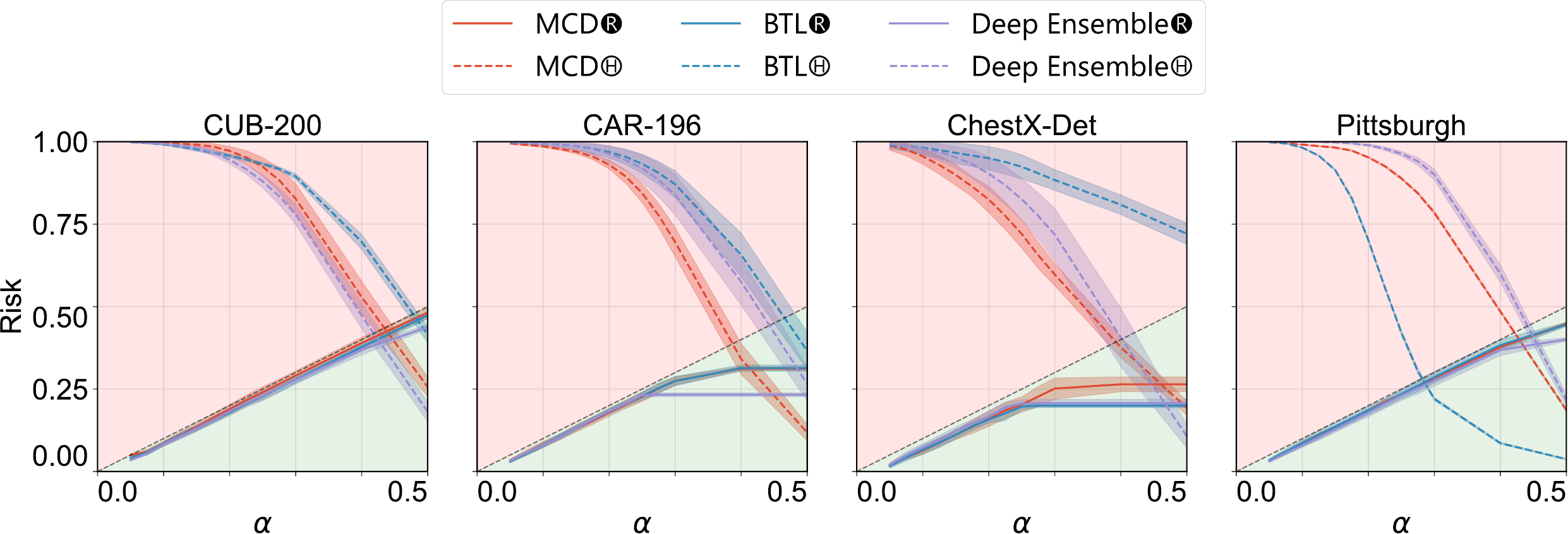} 
    \caption{The risks on different test sets: risk is effectively controlled by \sysname\ (see $\times$\raisebox{-0.03cm}{\sysr}) under different risk levels $\alpha$ (with $\delta=0.1$).}    
    \label{fig_multiple_risk_control}
\end{figure}

\begin{figure}[t]
    \centering
    \includegraphics[width=\columnwidth]{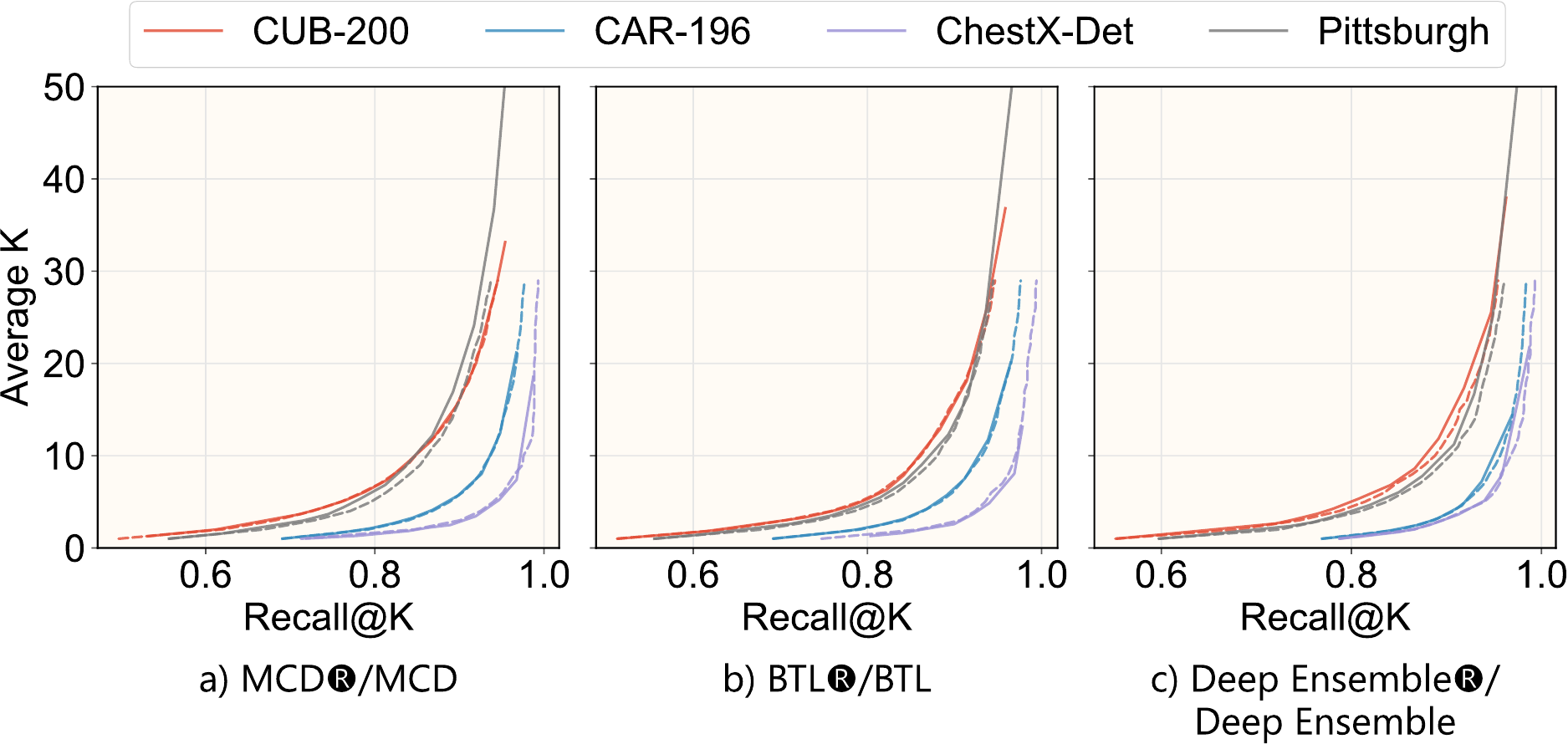} 
    \caption{The average $K$ across different methods on different test sets (\textit{solid line} $\rule[0.5ex]{0.15cm}{1pt}\rule[0.5ex]{0.15cm}{1pt}\rule[0.5ex]{0.15cm}{1pt}$  means $\times$\raisebox{-0.03cm}{\sysr} (with $\delta=0.1$), while \textit{dashed line} $\rule[0.5ex]{0.15cm}{1pt}~\rule[0.5ex]{0.15cm}{1pt}~\rule[0.5ex]{0.15cm}{1pt}$ means their conventional counterparts): When achieving the same recall, $\times$\raisebox{-0.03cm}{\sysr} and their fixed-size counterparts have similar average retrieval set sizes, indicating that $\times$\raisebox{-0.03cm}{\sysr} does not rely on simply increasing $K$ to control the risk.}
    \label{fig_multiple_k_control}
\end{figure}

\begin{figure}[h]
    \centering
    \includegraphics[width=\columnwidth]{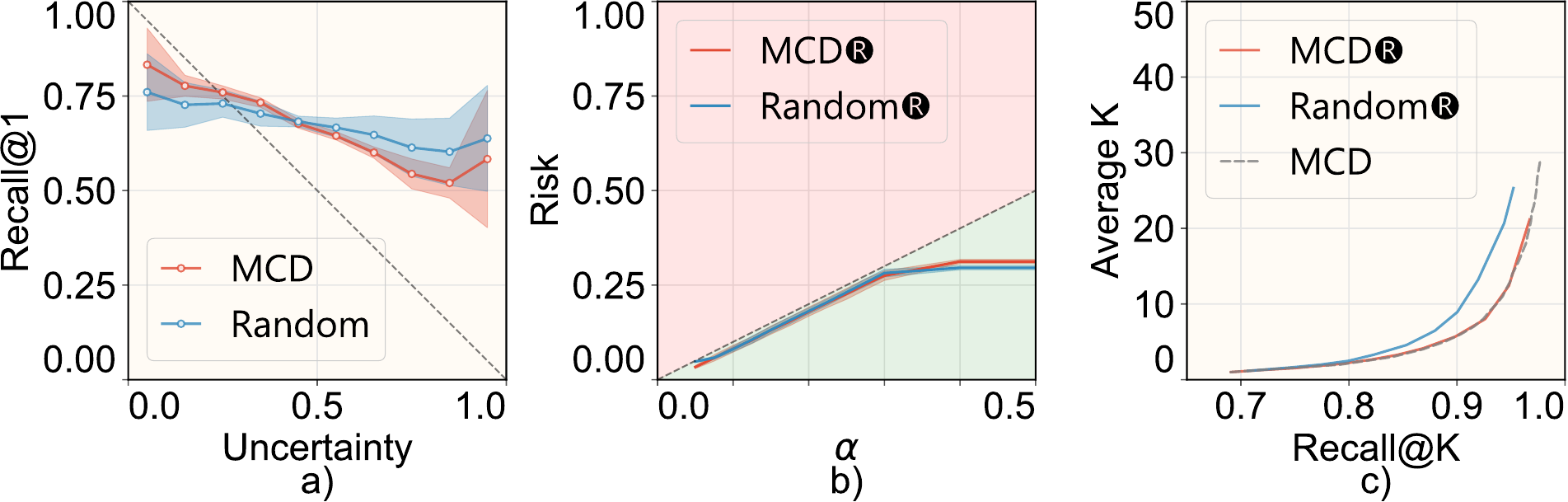} 
    \caption{The reliability diagram, risk and average $K$ of different models and pipelines on the CAR-196 dataset. Even with a poorly calibrated uncertainty (i.e., random guess), our $\times$\raisebox{-0.03cm}{\sysr\\} pipeline is still valid in controlling the risk. We also note that a poorly calibrated uncertainty results in a larger average retrieval set size.}
    \label{fig_risk_uncertainty_aba}
\end{figure}


    

\cref{fig_multiple_risk_control} shows the empirical risk of different image retrieval systems on different datasets. It is shown that the risks of $\times$\raisebox{-0.08\height}{\sysr} are always below the predefined risk levels $\alpha$.  
This means that before a retrieval we can say for sure that the retrievals of $\times$\raisebox{-0.08\height}{\sysr} have a $1-\alpha$ probability of covering the true nearest neighbors, with only a $\delta$ probability of being wrong. In comparison, the heuristic uncertainty-only methods, including MCD\sysh, BTL\sysh\ and Deep Ensemble\sysh, exhibit risks higher than the predefined risk levels in the range $\alpha<0.4$.  
The results show that the heuristic uncertainty alone is not reliable for risk control, while \sysname\ effectively controls the risk.

\begin{figure}[h]
    \centering
    \includegraphics[width=0.6\columnwidth]{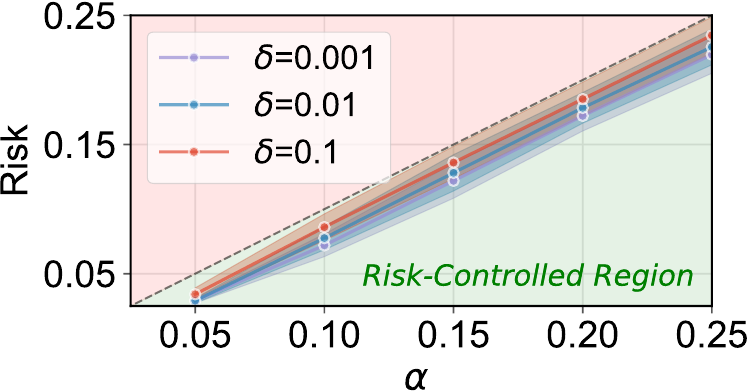} 
    \centering
    \caption{The risks on the Pittsburgh test set with different error rates by BTL\sysr: smaller $\delta$ leads to more conservative retrievals (e.g., larger retrieval size). }
    \label{fig_cub200_risk_control_delta}
\end{figure}

In addition, we investigate the effect of using different $\delta$. 
\cref{fig_cub200_risk_control_delta} shows the empirical risk with different $\delta$ on the Pittsburgh dataset. With a smaller $\delta$, \sysname\ would be more conservative (i.e., larger retrieval size) to ensure the risk is below the given $\alpha$. (see supplementary for similar trends on the other datasets.)

\subsubsection{Retrieval Set Size}

\sysname\ may always resort to large retrieval sets to lower the empirical risk, in which situation the risk is trivially controlled but the retrieval sets would be of less practical use. To examine this, we compare MCD\sysr\, BTL\sysr\ and Deep Ensemble\sysr\ against their fixed-size counterparts, MCD, BTL and Deep Ensemble. \cref{fig_multiple_k_control} shows that when achieving the same recall@$K$, $\times$\raisebox{-0.08\height}{\sysr} and their fixed-size counterparts have similar average retrieval set sizes. This indicates that $\times$\raisebox{-0.08\height}{\sysr} does not rely on simply increasing $K$ to control the risk. But we also notice that MCD\sysr\ and Deep Ensemble\sysr\ on the Pittsburgh test set have a slightly larger average retrieval set size than their fixed-size counterparts. This is due to poor uncertainty estimation on the Pittsburgh test set, as shown in \cref{fig_multiple_ece}.

\subsubsection{The impact of uncertainty estimator}

To study the impact of the uncertainty estimator on risk control performance, we modify the MCD image retrieval system by replacing the uncertainty estimator $f_u$ with a random uncertainty estimator. This random uncertainty estimator assigns a randomly sampled value from the uniform distribution $[0, 1]$ as the uncertainty for each query. Such uncertainty is poor as it does not correlate with the recognition performance. We refer to this modified image retrieval pipeline as \textbf{Random} and the corresponding RCIR system as \textbf{Random}\sysr.
\cref{fig_risk_uncertainty_aba}a) presents the reliability diagram of the uncertainty estimator of MCD and Random, where the curve of Random diverges more from the ideal calibration line (dashed line), indicating a poorly calibrated uncertainty. \cref{fig_risk_uncertainty_aba}b) shows the empirical risk of MCD\sysr\ and Random\sysr, while \cref{fig_risk_uncertainty_aba}c) displays the average $K$ of different pipelines. It can be seen that Random\sysr\ is still valid in controlling the risk, even with a poorly calibrated uncertainty. This is because the validity of the \cref{thm_ucb} is not dependent on the uncertainty estimator. However, using a poorly calibrated uncertainty estimator leads to a larger retrieval set size due to suboptimal computation of $\kappa$. This observation is consistent with the observation found by \cite{angelopoulos2021gentle}.

\subsubsection{Qualitative visualization} 
The distribution of retrieval size on the Pittsburgh test set is presented in \cref{fig_pitts_k_spread} (see the supplementary materials for similar results on the other datasets). It is clear that the retrieval size varies with the risk level $\alpha$: a smaller $\alpha$ results in a larger retrieval size, and vice versa. This helps end users save time on easy queries and focus on more difficult ones.

\begin{figure}[h]
    \centering
    \includegraphics[width=0.6\columnwidth]{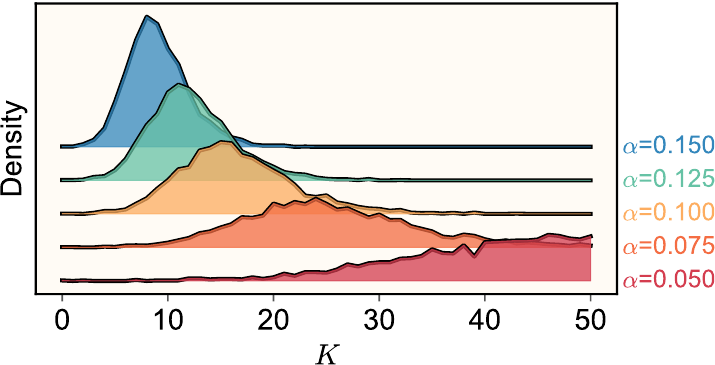} 
    \caption{The retrieval set size, $K$, on the Pittsburgh test set by BTL\sysr: retrieval set size adapts to the risk level $\alpha$ ($\delta=0.1$).} 
    \label{fig_pitts_k_spread}
\end{figure}

\begin{figure}[!t]
    \centering
    \includegraphics[width=\columnwidth]{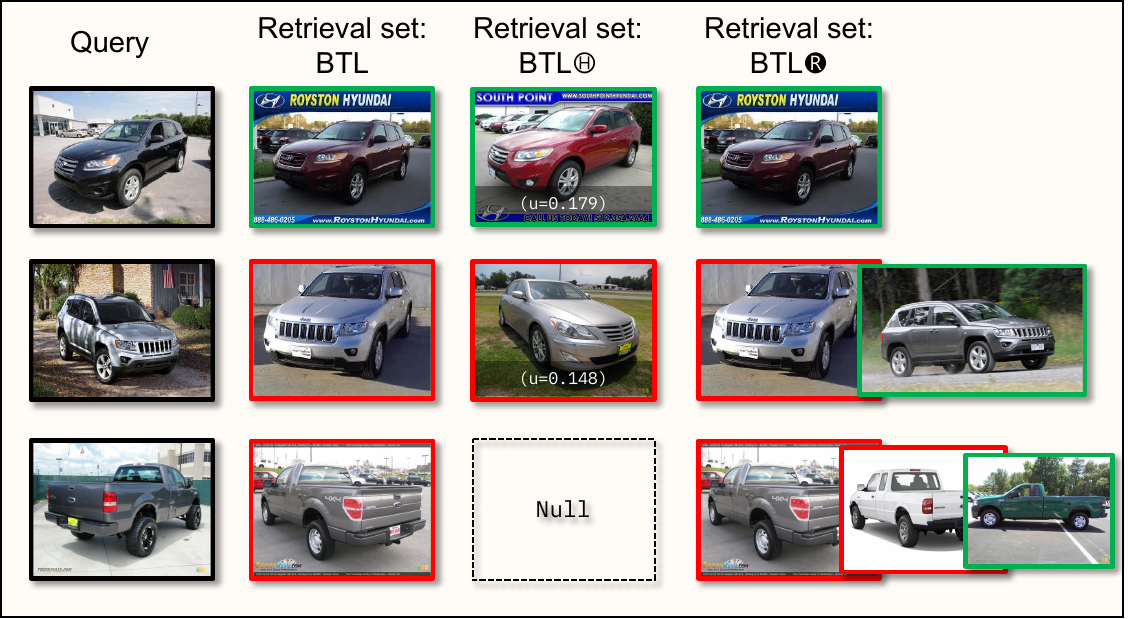} 
    \caption{The qualitative visualization of retrieval sets on the CAR-196 test set by different methods ({\textcolor{color_wrong}{Red border} \raisebox{-0.05cm}{\includegraphics[width=14pt]{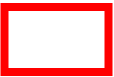}}} denotes a wrong candidate, while {\textcolor{color_correct}{green border} \raisebox{-0.05cm}{\includegraphics[width=14pt]{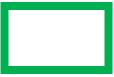}}} correct): BTL\sysr\ adjusts the size of retrievals to reflect the uncertainty while ensuring coverage ($\alpha=0.2$ and $\delta=0.1$).}
    \label{fig_qua}
\end{figure}
\cref{fig_qua} shows the qualitative visualization of retrievals on the CAR-196 dataset by different methods. In the first row, it is evident that when a relatively easy query is provided, all methods successfully retrieve the correct results. Moving on to the second row, where the query is more challenging, the fixed-size retrieval method fails without any indication, and BTL\sysh\ fails as well with a low heuristic number. However, BTL\sysr\ takes into account the difficulty of the query and retrieves an additional candidate, resulting in the correct answer. The third row represents a more demanding query. BTL\sysh\ generates no candidate, while BTL\sysr\ retrieves more candidates to meet the coverage requirement.
\section{Conclusion}
This paper proposes \sysname\ 
as a risk-controlled improvement to the current image retrieval systems, for the scenarios when the reliability of  image retrieval systems is essential. \sysname\ offers a coverage guarantee, regardless of data distribution or model selection.
Nevertheless, \sysname\ still has room for improvement. For example,  
while the choice of heuristic uncertainty estimator does not affect the validity of \sysname, a poorly calibrated uncertainty estimator may increase the average retrieval set size. It would be interesting to figure out a method that can not only quantify but also reduce the risks.  

\section*{Acknowledgements}
Xiaowei Huang's contribution is partially supported by the UK EPSRC through the End-to-End Conceptual Guarding of Neural Architectures [EP/T026995/1].
Xingyu Zhao's contribution is supported by the UK EPSRC New Investigator Award through Harnessing Synthetic Data Fidelity for Assured Perception of Autonomous Vehicles.
%
%
Special thanks to \url{https://instagram.com/littlemandyart} for the cute dog sticker pack design that is featured in \cref{fig_openfig}.

\bibliography{aaai25}

\end{document}